\documentclass[12pt]{article}
\usepackage{textcomp}
\usepackage{graphicx,psfrag,epsf}
\usepackage{enumerate}
\usepackage{natbib}
\usepackage{url} 
\usepackage{indentfirst}
\usepackage{amssymb}
\usepackage{algorithm}
\usepackage{algorithmic}
\usepackage{hyperref}
\usepackage{booktabs}
\usepackage{amsmath,soul}
\usepackage{graphicx}
\usepackage{float}
\usepackage{xcolor}
\usepackage{booktabs}
\usepackage{comment}
\usepackage{hyperref}
\usepackage[most]{tcolorbox}
\usepackage{longtable}

\usepackage{subfig}
\usepackage{graphicx}
\newcommand{\blind}{1}
\newcommand{\argmin}{\mathop{\mathrm{arg\,min}}}

\newtheorem{definition}{Definition}

\DeclareMathOperator{\arccosh}{arccosh}

\def\spacingset#1{\renewcommand{\baselinestretch}%
{#1}\small\normalsize} \spacingset{1}

\begin{document}

\def\spacingset#1{\renewcommand{\baselinestretch}%
{#1}\small\normalsize} \spacingset{1}


\if1\blind
{
  \title{\bf Automated Hierarchical Graph Construction for Multi-source Electronic Health Records}
   \author{Yinjie Wang$^1$\thanks{Wang and Zhou contributed equally to this work.}, Doudou Zhou$^2$\footnotemark[1], Yue Liu$^3$, Junwei Lu$^4$\thanks{Cai and Lu contributed equally as corresponding authors. Emails: \texttt{tcai@hsph.harvard.edu},  \texttt{junweilu@hsph.harvard.edu} }, Tianxi Cai$^4$\footnotemark[2] \\
   \small  $^1$Department of Statistics, University of Chicago \\
   \small  $^2$Department of Statistics and Data Science, National University of Singapore\\
    \small $^3$Department of Statistics, Harvard University\\
   \small  $^4$Department of Biostatistics, Harvard T.H. Chan School of Public Health 
    }
    \date{}
  \maketitle
} \fi

\if0\blind
{
  \bigskip
  \bigskip
  \bigskip
  \begin{center}
    {\LARGE\bf Automated Hierarchical Graph Construction for Multi-source Electronic Health Records}
\end{center}
  \medskip
} \fi
\date{}
\bigskip
\begin{abstract}
Electronic Health Records (EHRs), comprising diverse clinical data such as diagnoses, medications, and laboratory results, hold great promise for translational research.  EHR-derived data have advanced disease prevention, improved clinical trial recruitment, and generated real-world evidence. Synthesizing EHRs across institutions enables large-scale, generalizable studies that capture rare diseases and population diversity, but remains hindered by the heterogeneity of medical codes, institution-specific terminologies, and the absence of standardized data structures. These barriers limit the interpretability, comparability, and scalability of EHR-based analyses, underscoring the need for robust methods to harmonize and extract meaningful insights from distributed, heterogeneous data. To address this, we propose MASH (Multi-source Automated Structured Hierarchy), a fully automated framework that aligns medical codes across institutions using neural optimal transport and constructs hierarchical graphs with learned hyperbolic embeddings. During training, MASH integrates information from pre-trained language models, co-occurrence patterns, textual descriptions, and supervised labels to capture semantic and hierarchical relationships among medical concepts more effectively. Applied to real-world EHR data, including diagnosis, medication, and laboratory codes, MASH produces interpretable hierarchical graphs that facilitate the navigation and understanding of heterogeneous clinical data. Notably, it generates the first automated hierarchies for unstructured local laboratory codes, establishing foundational references for downstream applications.
\end{abstract}

\noindent%
{\it Keywords:} Hierarchical clustering, optimal transport, hyperbolic embeddings, medical coding systems, multi-source data. 
\vfill

\newpage
\spacingset{1.45} 

\section{Introduction}
\label{sec:introduction}

The widespread adoption of Electronic Health Record (EHR) systems has led to the accumulation of rich clinical data across healthcare institutions, creating substantial opportunities for predictive modeling \citep{lipton2015learning, rajkomar2018scalable}, decision support \citep{federico2015gnaeus, chunchu2012patient}, and knowledge discovery \citep{gu2021domain}. Despite this promise, the potential of EHR data remains underutilized in multi-institutional settings because of fundamental obstacles in how medical codes are adopted and organized \citep{hripcsak2013next, kush2008electronic}. These obstacles limit interoperability and interpretability, and they impede the portability of statistical and machine learning models across health systems.

The first major challenge in multi-institutional EHR analysis is the widespread use of local, institution-specific codes. Many health systems rely on internally defined identifiers for varying proportion of diagnoses, medications, and laboratory tests that do not map cleanly to standardized vocabularies such as the \emph{International Classification of Diseases} (ICD) for disease conditions \citep{world1978international, world2004international}, \emph{RxNorm} for standardized clinical drug nomenclature \citep{bennett2012utilizing}, and \emph{Logical Observation Identifiers Names and Codes} (LOINC) for laboratory tests \citep{mcdonald2003loinc}. These local codes are also often poorly documented, making them difficult to interpret. As a result, the same clinical concept, such as the C-reactive protein laboratory test, may be represented differently across institutions, creating substantial barriers to cross-system analysis and model generalizability. Overcoming this heterogeneity requires automated, data-driven \emph{code harmonization} methods that align semantically equivalent concepts across institutions into a shared representation.

The second major challenge arises from the excessive granularity and lack of unified hierarchical structure in EHR coding systems, which hinders concept-level aggregation and interpretability in clinical research. Standardized terminologies like ICD-10-CM and LOINC contain  tens of thousands of codes, many differing only in minor details such as laterality or specimen type. This leads to artificial fragmentation that complicates generalizable analysis \citep{wu2019mapping,mcdonald2003loinc}. Moreover, widely used systems often lack a unified and clinically coherent hierarchy; for example \emph{RxNorm} includes multiple overlapping hierarchies curated by different agencies \citep{pathak2010analyzing}, including resources within the \emph{Unified Medical Language System} (UMLS) \citep{bodenreider2004unified}. Manual curation of hierarchical mappings is labor-intensive, error-prone, and unsustainable as code systems grow. These limitations highlight the need for automated, data-driven \emph{hierarchical clustering} methods that can infer clinically meaningful structure, recover latent parent concepts, and organize codes into interpretable multi-level taxonomies to support modeling, cohort definition, and other downstream applications.

To address these challenges, a growing body of research has focused on learning low-dimensional embeddings of medical codes using empirical usage patterns and existing knowledge graphs  \cite{chen2024spectral,hou2014discriminative,li2016low,hong2021clinical}. For example, simple {\em word2vec}-based methods leverage co-occurrence statistics to derive embeddings that reflect functional relationships between codes \citep{hong2021clinical}. Others incorporate semantic information of the code descriptions by leveraging language models trained on biomedical corpora\citep{yuan2022coder, zeng2022automatic}, or integrate both co-occurrence statistics and semantic embeddings to improve robustness across domains\citep{zhou2022multiview, zhou2025representation}. While 
these methods provide strong foundations, embeddings trained independently across  institutions often reside in non-aligned geometric spaces, limiting their utility for multi-institutional analysis. Simple ad hoc alignment strategies, such as Procrustes analysis \citep{kementchedjhieva-etal-2018-generalizing}, are generally insufficient to achieve robust harmonization when coding practices differ substantially. More recent work has explored the use of optimal transport as a principled approach to cross-domain alignment \citep{perrot2016mapping, hoyos2020aligning}, though its adaptation to multi-institutional EHR data remain underdeveloped.

A related line of related research focuses on learning hierarchical representation of medical concepts. Curated resources such as the UMLS, RxNorm, and LOINC provide manually defined hierarchies that group concepts into clinically meaningful categories. While invaluable, these hierarchies are costly to maintain, inconsistent across domains, and typically exclude  the large number of institution-specific local codes common in real-world EHRs. To address these gaps, statistical and machine learning methods have been proposed to infer hierarchies directly from data. Latent tree models recover tree structures from observed similarities \citep{choi2011learning}, and hyperbolic embeddings exploit negatively curved geometry to represent hierarchical relationships efficiently \citep{krioukov2010hyperbolic, nickel2017poincare}. These methods have been shown success in areas such as knowledge graph completion \citep{nickel2015review, yao2019kg} and ontology alignment \citep{he2022bertmap}, demonstrating the promise of automated hierarchy construction. Recent work has used large language models (LLMs) to enrich biomedical ontologies and improve semantic interpretability \citep{yuan2022coder, zeng2022automatic}. Nevertheless, most existing approaches assume relatively clean vocabularies and struggle with the noisy, heterogeneous, and institution-specific coding systems found in practice. There remains a critical need for automated methods that can jointly harmonize codes across institutions and construct clinically coherent, interpretable hierarchies at scale.

In this paper we introduce \textbf{MASH} (\textbf{M}ulti-source \textbf{A}utomated \textbf{S}tructured \textbf{H}ierarchy), a unified framework designed to address the dual challenges of code harmonization and hierarchy construction in large-scale EHR systems. We demonstrate its effectiveness using data from two major U.S. health systems, the \emph{U.S. Department of Veterans Affairs} (VA) and \emph{Mass General Brigham} (MGB), though the methods developed are broadly generalizable and applicable to a wide range of multi-institutional healthcare settings. Despite sharing many clinical goals, the VA and MGB differ substantially in their coding conventions and clinical workflows. Most MGB codes are mapped to standard ontologies with established hierarchical structures, whereas a large portion of VA codes remain unmapped and lack any curated hierarchy, making alignment and organization particularly challenging. MASH addresses this by integrating co-occurrence statistics with semantic signals from biomedical language models, aligning embeddings across institutions using neural optimal transport, and constructing clinically coherent hierarchies through hyperbolic embedding and recursive clustering, The framework is guided by partial supervision from existing ontologies and further enhanced by LLM-generated annotations to improve usability and interpretability.

Applied to VA and MGB data, which collectively represent millions of patients and diverse coding practices, MASH delivers two key outcomes. First, it harmonizes institution-specific codes into a common representational space, enabling previously infeasible cross-system analyses. Second, it produces hierarchical graphs for diagnosis, medication, and laboratory tests, including the first automated hierarchies for the extensive set of VA local laboratory codes that lack standardized references. In doing so, MASH not only reconstructs known structures such as PheCode and RxNorm hierarchies but also generates clinically meaningful organization for previously unstructured codes. This case study demonstrates how automated, data-driven approaches can resolve long-standing barriers to multi-institutional EHR research and lay the groundwork for interpretable and portable analytics.

The remainder of the paper is organized as follows. Section~\ref{sec:data} describes the VA and MGB datasets and preprocessing steps. Section~\ref{sec:method} details the framework for harmonization and hierarchy construction. Section~\ref{sec:realdata} presents the results for VA and MGB codes. Section~\ref{sec:discussion} concludes with broader implications and directions for future research, particularly in federated EHR environments. Additional numeric results are presented in Supplementary Material.

\section{Data Sources and Representation Construction}
\label{sec:data}
Our study focuses on structured EHR data from the VA and MGB, covering over $15$ million patients and capturing highly diverse coding practices, making them a natural case study for the challenges of code harmonization and hierarchy construction. The VA dataset includes records for 12.6 million patients from 1999 to 2019 across more than 150 facilities, while the MGB dataset comprises 2.5 million patients from 1998 to 2018. From each system, we extracted three key categories of structured data: diagnosis codes, medication prescriptions, and laboratory test results. Procedure codes, which often involve CPT coding, were excluded from our analysis due to licensing restrictions that prevent the construction and distribution of derived hierarchies.

We organized structured EHR data across institutions using existing hierarchical ontologies: PheCode for diagnoses, RxNorm for medications, and LOINC for laboratory tests. For diagnosis data, ICD codes were mapped to clinically meaningful phenotypes using the PheCode ontology \citep{phewascatalog}. Medication data were standardized to RxNorm; while most VA medication records already use RxNorm identifiers, 199 local codes required additional mapping, which we performed by abstracting away dosage-specific details. Laboratory codes at MGB are mapped to LOINC, which provides hierarchical organization of laboratory concepts. In contrast, the VA dataset includes a large number of local laboratory codes, only $17\%$ of which are mapped to LOINC or higher-level laboratory categories. To incorporate broader laboratory concepts, we used the LOINC Multiaxial framework \citep{mcdonald2004logical}, which groups related lab tests under higher-level LOINC Part (LP) codes.

To improve semantic clarity and reduce redundancy, we consolidated local laboratory codes with identical text descriptions. We further excluded rare codes, defined as those occurring fewer than $5,000$ times in the VA data or $1,000$ times in the MGB data, following standard practice \citep{hong2021clinical}. After preprocessing, the resulting datasets contained $5,961$ codes for MGB and $5,560$ for VA, with $2,985$ codes appearing in both systems (Table~\ref{table_overlap_unique_counts}). These figures highlight the integration challenge: while some code overlap exists, both systems, especially the VA, retain many unique codes, with the VA local laboratory codes lacking standardized ontology support.

To support MASH algorithm training, we assembled three complementary sources of information. First, empirical co-occurrence matrices, derived as summary data from EHR, reflect how codes are used together in clinical practice, capturing implicit workflow and contextual associations. Second, textual descriptions of the EHR codes offer semantic information that is particularly valuable for rare or unmapped codes. Third, existing curated hierarchies provide structural supervision where available. Combined, these sources allow us to embed codes in a unified space and construct clinically coherent hierarchies. We describe each in detail below.

\begin{table}[ht]
\centering
\begin{tabular}{lccccc}
\toprule
              & \multicolumn{1}{c}{Diagnoses} &
                \multicolumn{1}{c}{Medications} &
                \multicolumn{2}{c}{Laboratory Tests} &
                \multicolumn{1}{c}{Total} \\
\cmidrule(lr){2-2}\cmidrule(lr){3-3}\cmidrule(lr){4-5}\cmidrule(lr){6-6}
              & PheCodes & RxNorm & LOINC & Local Lab &  \\ 
\midrule
MGB only     & 79   & 413  & 2484 & 0    & 2976 \\
Overlap      & 1715 & 1081 & 189  & 0    & 2985 \\
VA only      & 61   & 468  & 193  & 1853 & 2575 \\
\bottomrule
\end{tabular}%
\caption{Counts of overlapping and unique PheCode, RxNorm, LOINC, and local laboratory codes in the MGB and VA datasets.}
\label{table_overlap_unique_counts}
\end{table}

\subsection{Co-occurrence Statistics}

Co-occurrence patterns capture how codes appear together in patient records, providing a strong empirical signal for learning embeddings via the matrix factorization variant of the skip-gram algorithm \citep{mikolov2013distributed,goldberg2014word2vec,beam2020clinical,hong2021clinical}. Frequently co-occurring codes within short time windows are often clinically related, even in the absence of explicit ontology links. As aggregated and de-identified summaries, co-occurrence matrices also enable cross-institutional analysis without exposing patient-level data. For LOINC Part codes, which are not directly recorded in EHRs, we computed co-occurrence by aggregating over their descendant LOINC codes. Following \citet{hong2021clinical}, we constructed Shifted Positive Pointwise Mutual Information (SPPMI) matrices, where $\mathcal{C}_k(i,j)$ denotes the total co-occurrence of codes $i$ and $j$ within 30-day windows for institution $k$. The $(i,j)$th entry of the SPPMI matrix is defined as:
\begin{align}
\label{defsppmi}
\mathbb{SPPMI}_k(i, j) = \max \left\{0, \log \frac{\mathcal{C}_k(i, j)}{\mathcal{C}_k(i, \cdot)\,\mathcal{C}_k(j, \cdot)} \right\},
\end{align}
where $\mathcal{C}_k(i, \cdot) = \sum_{i} \mathcal{C}_k(i, k)$ denotes the marginal totals. For each institution, we decomposed the SPPMI matrix using singular value decomposition (SVD) and extracted low-dimensional embeddings, followed by $\ell_2$ row normalization.

\subsection{Semantic Information}

Textual descriptions of medical codes offer expert-defined semantics that complement co-occurrence statistics. Codes that may not frequently co-occur, especially across institutions, often share descriptive similarity. To exploit this information, we used the CODER language model \citep{yuan2022coder, zeng2022automatic}, trained on biomedical concept pairs extracted from UMLS via contrastive learning. CODER places semantically similar codes closer in the embedding space. For each code, we obtained a CODER-based embedding from its textual description and normalized it to unit length. Codes were thus associated with both semantic embeddings and institution-specific co-occurrence embeddings, ensuring that each representation captured both usage patterns and semantic meaning.

\subsection{Hierarchical Information}
\label{hierinfo}

When available, curated hierarchies serve as valuable supervision signals and benchmarks. They encode parent–child relations that guide embeddings toward clinically meaningful structure. For diagnoses, hierarchical information is defined within the PheCode system. For medications, we relied on the VA classification system through NDF-RT \citep{nelson2011normalized}. For laboratory codes, we used the LOINC Multiaxial framework \citep{mcdonald2004logical}. VA local laboratory codes, in contrast, lack any inherent hierarchy. To evaluate the hierarchies constructed for these codes, we leveraged available mappings from LOINC to local laboratory codes to impute hierarchical relations.

We also assembled labeled pairs of codes, distinguishing between \emph{similar} codes (for example synonyms or fine-grained variants) and \emph{related} codes (for example functional or ontological links), using curated resources such as the UMLS Metathesaurus \citep{bodenreider2004unified}. These labeled pairs, covering PheCodes, RxNorm, laboratory codes, and LOINC–local lab links, were split evenly into training and testing sets for supervision and evaluation.

\section{Method}
\label{sec:method}

\begin{figure}[t!]
    \centering
    \includegraphics[width=1\textwidth]{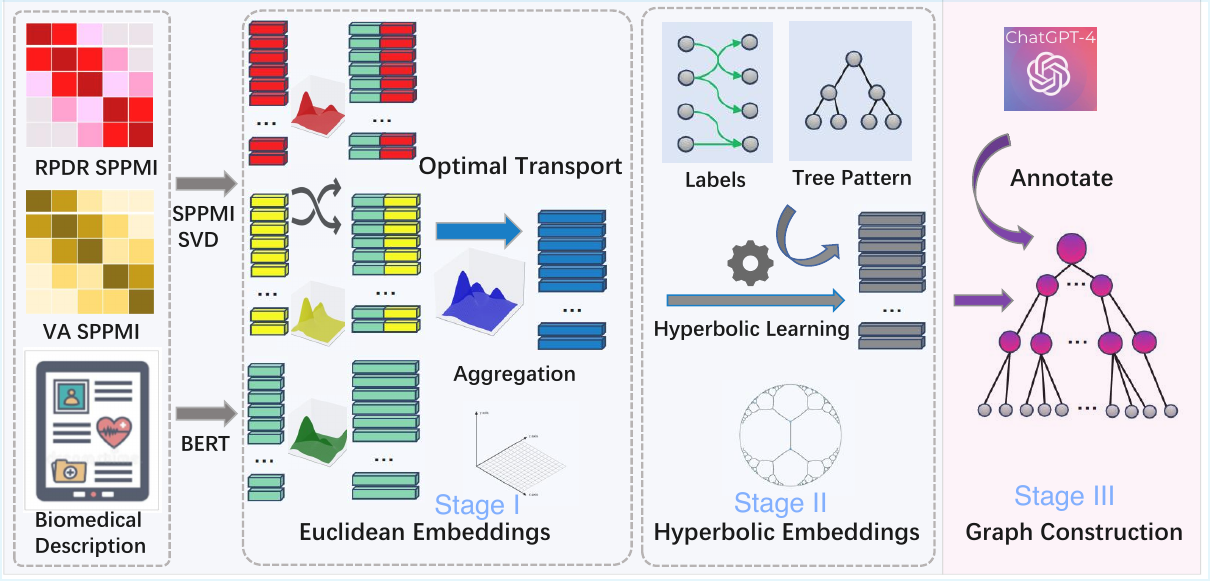}
      \caption{Pipeline of MASH in three stages: (I) Embeddings from co-occurrence and semantics are aligned across institutions via neural optimal transport; (II) these embeddings are mapped into hyperbolic space to capture hierarchy; and (III) a recursive grouping procedure constructs hierarchical graphs. Large language models provide interpretable annotations for latent nodes.}
    \label{workflow}
\end{figure}

MASH provides an automated framework that harmonizes local and system-specific codes and constructs clinically meaningful hierarchies in the VA and MGB datasets through a three-stage process. In Stage I, embeddings derived from co-occurrence statistics and textual semantics are aligned across institutions using a neural optimal transport procedure, producing a shared representation that unifies comparable codes across systems. In Stage II, the harmonized embeddings are mapped into hyperbolic space, which naturally models hierarchical structures, with optimization guided by partial supervision and an additivity loss enforcing consistency with tree structures. In Stage III, a recursive grouping strategy reconstructs the code hierarchy from these embeddings, introducing latent nodes as needed. We deploy large language models to annotate the resulting clusters to enhance  interpretability. Figure~\ref{workflow} provides an overview of the pipeline.

\subsection{Stage I: Aggregation Through Optimal Transport}
\label{subsecot}

In Stage I, we use optimal transport (OT) to align heterogeneous SPPMI-SVD embedding spaces that differ in scale and geometry across EHR systems. We initialize the representation of each code as a $d$-dimensional vector formed by concatenating a $d/2$-dimensional CODER embedding of the code description with a $d/2$-dimensional SPPMI-SVD embedding ($d/2=768$) derived from EHR. Let $\mathcal{E}_{\text{MGB}}=\{e_c^{\text{MGB}}\}$ and $\mathcal{E}_{\text{VA}}=\{e_c^{\text{VA}}\}$. A neural OT mapping $T_1$ is trained so that $T_1(e_c^{\text{VA}})\approx e_c^{\text{MGB}}$, and a second mapping $T_2$ aligns $\mathcal{E}{\text{CODER}}$ to the MGB space. The aligned sets $T_1(\mathcal{E}_{\text{VA}})$, $T_2(\mathcal{E}_{\text{CODER}})$, and $\mathcal{E}_{\text{MGB}}$ are averaged, concatenated with CODER embeddings, and row-wise $\ell_2$ normalization is applied to yield the harmonized representation.

Formally, let $\mathcal{E}_{\mathcal{S}}$ and $\mathcal{E}_{\mathcal{T}}$ denote the source and target embedding spaces. For the set of codes shared between source and target domains, we extract corresponding embeddings $\{e_i^s\}_{i=1}^{m} \subset \mathcal{E}_{\mathcal{S}}$ and $\{e_i^t\}_{i=1}^{m} \subset \mathcal{E}_{\mathcal{T}}$. Define the matrices
\[
E_s = [e_1^s, \dots, e_{m}^s]^T, \quad E_t = [e_1^t, \dots, e_{m}^t]^T.
\]
The goal is to learn a neural transformation $T$ and a coupling matrix $\pi$ that minimize the cost of transporting $E_s$ into $E_t$:
\begin{equation}
    \widehat{\pi}, \widehat{T} = \argmin_{\pi \in \Pi,\, T} \int c(T(e_s), e_t) \, d\pi(e_s, e_t),
    \label{eq:int}
\end{equation}
where $c(\cdot, \cdot)$ denotes the squared Euclidean distance and $\pi(e_s, e_t) = \pi_{st}$. The feasible set $\Pi$ includes all doubly stochastic matrices:
\[
\Pi := \left\{ \pi \in (\mathbb{R}^+)^{m \times m} \;\middle|\; \pi \mathbf{1}_m = \frac{1}{m} \mathbf{1}_m = \pi^T \mathbf{1}_m  \right\}, \ \mbox{where $\mathbf{1}_m$ is an $m\times 1$ vector of ones.}
\]
 The transport function $T$ is parameterized by a neural network with three hidden layers of sizes $8000$, $12000$, and $8000$, each followed by a ReLU activation function, enabling a flexible nonlinear mapping between embedding spaces.

To make this problem computationally tractable, the integral formulation \eqref{eq:int} can be rewritten in terms of barycentric projections \citep{peyre2019computational}. For a given transport plan $\pi$, each source embedding $e_i^s$ is associated with a barycentric correspondence in the target space,
\[
B_{\pi,i} = \frac{\sum_{j=1}^m \pi_{ij} e_j^t}{\sum_{j=1}^m \pi_{ij}},
\]
which represents the weighted average of target embeddings matched to $e_i^s$. The transformation $T$ is then trained so that $T(e_i^s)$ is close to $B_{\pi,i}$, aligning transported source codes with their barycentric targets. This reformulation naturally decomposes the OT objective into a \emph{mapping loss}, capturing the discrepancy between $T(E_s)$ and $B_\pi$, and an \emph{OT loss}, capturing the transport cost implied by $\pi$.

We jointly learn $T$ and $\pi$ by minimizing the following objective \citep{perrot2016mapping}:
\begin{align}
\label{OTtotallloss}
    \mathcal{L}_{OT} = \underbrace{\| T(E_s) - B_{\pi} \|^2_{\mathcal{F}}}_{l_0(T, B_{\pi})\,\,\text{(mapping loss)}} 
    + 
    \underbrace{\eta \langle \pi, C_T \rangle_{\mathcal{F}}}_{\text{OT loss}},
\end{align}
where $T(E_s)$ is the transformed source embedding matrix, and $B_\pi$ is the barycentric projection defined above, with the $i$th row $B_{\pi,i}$. The cost matrix $C_T \in \mathbb{R}^{m \times m}$ contains squared distances between $T(e_i^s)$ and $e_j^t$, and $\eta$ is a hyperparameter balancing the two loss terms.

Optimizing \eqref{OTtotallloss} requires updating both the neural mapping $T$ and the coupling $\pi$. Because the objective is not jointly convex in $(T,\pi)$, we adopt a block coordinate descent (BCD) strategy \citep{tseng2001convergence, perrot2016mapping, hoyos2020aligning}, which alternates between (i) fixing $\pi$ and updating $T$ via stochastic gradient descent (SGD), and (ii) fixing $T$ and updating $\pi$ via a linear program. In each inner loop, $\pi^*$ is obtained from the linear oracle, and the step size $\alpha^k$ is chosen according to the Armijo rule \citep{armijo1966minimization}. This alternating scheme is widely used in OT-based embedding alignment and provides practical convergence guarantees. Algorithm~\ref{otoptimize} summarizes the procedure.

\begin{algorithm}[h!]
\caption{Block Coordinate Descent for Optimal Transport}
\label{otoptimize}
\begin{algorithmic}
    \STATE \textbf{Input:} Initial coupling $\pi^0 = \frac{1}{m} \mathbf{I}_m$; Initial mapping $T^0$ (neural network); learning rate $r = 1e{-}3$; hyperparameters $\omega = 1e{-}4$, $\eta = 1e{-}5$; number of outer loops $M = 2$; inner loop limits $N_1 = 1000$ and $N_2 = 50$.
    \FOR{$k = 0$ to $M$}
        \STATE \textbf{Update $T^k$ with fixed $\pi^k$}: \\
        \FOR{epoch $\le N_1$}
            \STATE Minimize $l_0(T, \pi^k)$ using SGD with learning rate $r$ to obtain $T^{k}$.
        \ENDFOR
        \STATE $T^{k+1} = T^k$
        \STATE \textbf{Update $\pi^{k}$ with fixed $T^{k + 1}$}: \\
        \STATE Compute cost matrix $C_{T^{k+1}}$. 
        \FOR{iteration $\le N_2$}
            \STATE Solve linear program to get $\pi^{*} = \argmin_{\pi \in \Pi} \langle \pi, \eta C_{T^{k+1}} + \nabla_{\pi} l_0(T^{k+1}, B_{\pi})\mid_{\pi = \pi^k}\rangle$. 
            \STATE Update $\pi^k = \pi^k + \omega \alpha^k (\pi^* - \pi^k)$, with $\alpha^k$ selected by the Armijo rule.
        \ENDFOR
        \STATE $\pi^{k+1} = \pi^k$
    \ENDFOR
\end{algorithmic}
\end{algorithm}

\subsection{Stage II: Learn Embeddings in Hyperbolic Space}
\label{subsehyplearn}

In Stage II, we map the harmonized embeddings into hyperbolic space to capture hierarchical structure, leveraging the exponential growth of distances to model tree-like code hierarchies \citep{nickel2017poincare}. Hyperbolic space is defined as the set of points $\mathbf{z}=(z_0,z_1,\ldots,z_d)^\top \in \mathbb{R}^{d+1}$ with $z_0>0$, equipped with the Lorentzian inner product
\[
\langle \mathbf{z}, \mathbf{z}' \rangle_{\mathcal{L}} = -z_0 z_0' + \sum_{i=1}^{d} z_i z_i'.
\]
The $d$-dimensional hyperbolic space is
\[
\mathbb{H}^d = \left\{ \mathbf{z} \in \mathbb{R}^{d+1} \;\middle|\; \langle \mathbf{z}, \mathbf{z} \rangle_{\mathcal{L}} = -1,\; z_0 > 0 \right\},
\]
and the geodesic distance between $\mathbf{z}, \mathbf{z}' \in \mathbb{H}^d$ is
$d_{\ell}(\mathbf{z}, \mathbf{z}') = \arccosh\!\left(-\langle \mathbf{z}, \mathbf{z}' \rangle_{\mathcal{L}}\right).$
Hyperbolic geometry is well suited for hierarchies because volume grows exponentially with radius, enabling deep structures to be embedded in low dimensions without the distortion that would require prohibitively high Euclidean dimensions \citep{krioukov2010hyperbolic}; thus, hyperbolic embeddings serve as a compact continuous analogue of trees.

In our framework, observed medical codes serve as leaf nodes, and in some cases, as penultimate nodes of an ideal latent tree. Higher-level abstract concepts, which are not directly observed in the data, are represented as \textbf{latent nodes}. To recover these, we employ a tree-structured graphical model \citep{cowell1999probabilistic}, where distances are computed in hyperbolic space. Leveraging the additivity property of information distances in trees \citep{erdos1999few}, we assume:
\begin{align}
\label{addequation}
d_{\ell}(\mathbf{z}_k, \mathbf{z}_l) = \sum_{(i, j) \in \text{Path}(k, l)} d_{\ell}(\mathbf{z}_i, \mathbf{z}_j),
\end{align}
where $\text{Path}(k, l)$ denotes the shortest path between $\mathbf{z}_k$ and $\mathbf{z}_l$, possibly traversing latent, unobserved nodes. This additivity principle underpins our hierarchy recovery approach; further details are provided in Supplementary Section~\ref{additivity condition}.

To enforce additivity, we require that if $\mathbf{z}_i$ is the parent of $\mathbf{z}_j$ and $\mathbf{z}_k$ is not a descendant of $\mathbf{z}_j$, then  
\begin{align}
\label{addsuffequation}
d_{\ell}(\mathbf{z}_j, \mathbf{z}_k) = d_{\ell}(\mathbf{z}_i, \mathbf{z}_k) + d_{\ell}(\mathbf{z}_i, \mathbf{z}_j),
\end{align}
which ensures the additivity property in \eqref{addequation}. We define the \emph{additivity loss} as  
\[
\mathcal{L}_a = \frac{1}{|I_{pa}|} \sum_{(i, j, k) \in I_{pa}} \left( d_{\ell}(\mathbf{z}_j, \mathbf{z}_k) - d_{\ell}(\mathbf{z}_i, \mathbf{z}_k) - d_{\ell}(\mathbf{z}_i, \mathbf{z}_j) \right)^2,
\]
where $I_{pa} = \{(i,j,k) \mid i \text{ is parent of } j,\; k \text{ not descendant of } j\}$. During training, $I_{pa}$ is constructed from known or inferred hierarchies.

To initialize the hyperbolic embeddings, we project the $\ell_2$-normalized Euclidean embeddings into $\mathbb{H}^d$ via
\[
f: \mathbb{R}^d \to \mathbb{H}^d, \quad f(x_1, \dots, x_d) = (\sqrt{2}, x_1, \dots, x_d),
\]
which is valid since all Euclidean vectors satisfy $\|x\|_2 = 1$. To preserve the geometry of the original embedding space, we introduce an \emph{information-preserving loss}:
\[
\mathcal{L}_e = \frac{1}{n} \sum_{i,j} \Big( \langle \mathbf{z}_i, \mathbf{z}_j \rangle_{\mathcal{L}} - I_{ij} \Big)^2,
\]
where $I_{ij}$ denotes the Lorentzian product between the initialized hyperbolic embeddings of codes $i$ and $j$, and $n$ is the number of unique codes across the two health systems.

We further incorporate supervision through a contrastive learning framework \citep{oord2018representation, wang2019multi}. 
Contrastive learning refines the embedding space by pulling similar or related codes closer together while pushing dissimilar codes apart. 
This promotes discriminative structure and supports downstream tasks such as classification and hierarchy construction. 
In particular, we adopt the InfoNCE contrastive loss \citep{oord2018representation}, which is well-suited for learning robust representations because it effectively balances positive and negative pairs and provides stable gradients even with large vocabularies:
\[
\mathcal{L}_c = -\frac{1}{|I_p|} \sum_{(i, j) \in I_p} 
\log \frac{e^{-d_{\ell}(\mathbf{z}_i, \mathbf{z}_j)}}{\sum_{k: (i, k) \notin I_p } e^{-d_{\ell}(\mathbf{z}_i, \mathbf{z}_k)} },
\]
where $I_p$ denotes the index set of positively labeled code pairs. 
This loss encourages embeddings of codes in $I_p$ to be close in hyperbolic space while contrasting them against unrelated pairs sampled from the vocabulary.

During training, we first initialize Euclidean embeddings in hyperbolic space, then optimize a weighted combination of losses:
\[
w_a \mathcal{L}_a + w_e \mathcal{L}_e + w_c \mathcal{L}_c,
\]
where $w_a$, $w_e$, and $w_c$ are hyperparameters controlling the contributions of the additivity, information-preserving, and contrastive terms. The resulting embeddings integrate co-occurrence patterns, geometric consistency, and supervised semantic relationships, forming a strong foundation for hierarchy construction.

\subsection{Stage III: Graph Construction}
\label{subsehiercluster}

After obtaining hyperbolic embeddings that integrate multi-source information and approximately satisfy the additivity property, we leverage \eqref{addequation} to reconstruct the hierarchical tree including latent nodes from the embeddings of observed medical codes. We apply recursive grouping with $k$-means clustering \citep{choi2011learning} to iteratively recover tree structure. Before describing the algorithm, we introduce the following definition:

\begin{definition}
Given a tree with node set $\mathcal{V}$, a subset of nodes $\mathcal{B} \subset \mathcal{V}$ is called a \textbf{bottom set} if every pair of nodes in $\mathcal{B}$ are either siblings or in a parent-child relationship, and all non-parent nodes in $\mathcal{B}$ are leaf nodes.
\end{definition}

At each iteration of recursive grouping, we identify a bottom set $\mathcal{B}$ within the current node set $\mathcal{V}$. A latent node is then introduced as the common ancestor of the nodes in $\mathcal{B}$, establishing new parent–child relationships. The elements in $\mathcal{B}$ are replaced by this latent node, resulting in an updated node set.

\begin{figure}[H]
    \centering
    \includegraphics[width=0.7\textwidth]{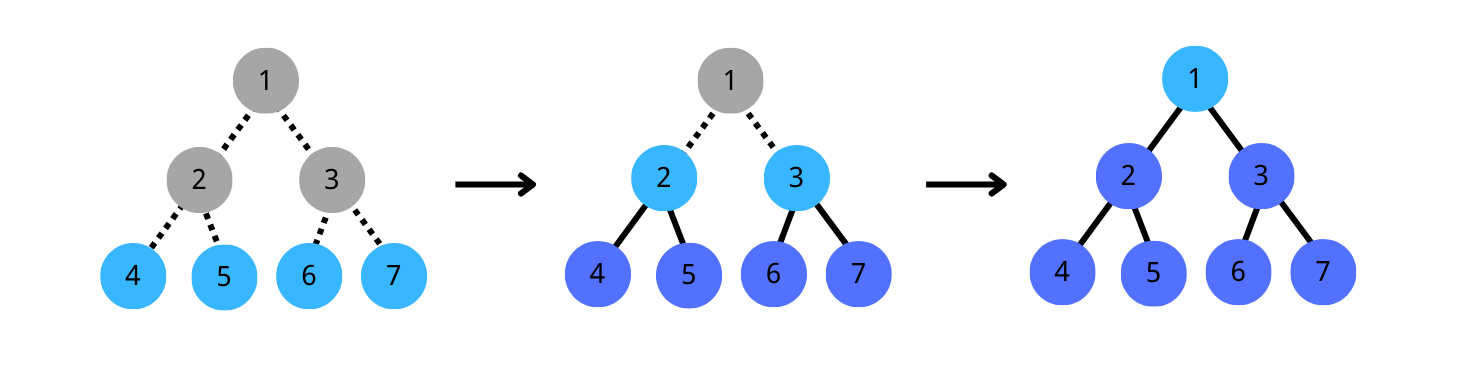}
      \caption{Recursive grouping algorithm example. Dashed lines and grey nodes represent latent structure to be discovered. Light blue nodes indicate the current node set $\mathcal{V}$ at each iteration. Solid lines and dark blue nodes represent recovered structure.}
    \label{RGprocesspic}
\end{figure}

Figure~\ref{RGprocesspic} illustrates this process. Initially, $\mathcal{V} = \{4, 5, 6, 7\}$. Two bottom sets $\{4, 5\}$ and $\{6, 7\}$ are identified, leading to the introduction of latent parent nodes $2$ and $3$. The updated node set becomes $\{2, 3\}$. A new distance matrix is computed using \eqref{addequation}, and the procedure repeats. In the second iteration, nodes $2$ and $3$ are grouped under a final latent root node, completing the tree.

The key challenge lies in identifying bottom sets from  embeddings. According to \eqref{addequation}, if nodes $i$ and $j$ belong to the same bottom set, the difference $d_{\ell}(\mathbf{z}_i, \mathbf{z}_k) - d_{\ell}(\mathbf{z}_j, \mathbf{z}_k)$ remains approximately constant for all other nodes $k$. This motivates the following criterion:
\begin{equation}
    D_{ij} := \max_{ k \in \mathcal{V} \setminus \{i, j\} } \left( d_{\ell}(\mathbf{z}_i, \mathbf{z}_k) - d_{\ell}(\mathbf{z}_j, \mathbf{z}_k) \right) - \min_{ k \in \mathcal{V} \setminus \{i, j\} } \left( d_{\ell}(\mathbf{z}_i, \mathbf{z}_k) - d_{\ell}(\mathbf{z}_j, \mathbf{z}_k) \right).
\label{Dij}
\end{equation}
If \eqref{addequation} is satisfied, nodes $i$ and $j$ belong to the same bottom set if and only if $D_{ij} = 0$. A more detailed discussion about this criterion can be found in Supplementary Section~\ref{subsec:Dij}. 

In practice, due to noise and estimation error, $D_{ij}$ will not be exactly zero. We therefore compute the matrix $[D_{ij}]_{1 \le i,j \le n(\mathcal{V})}$ and apply $k$-means clustering to group nodes into bottom sets. The number of clusters is chosen via the silhouette method \citep{schwarz1978estimating}.

After identifying bottom sets $\{\mathcal{B}_i\}_{1 \le i \le n_B}$, we assign a latent node $h_i$ to each $\mathcal{B}_i$. If a parent–child relationship (Section \ref{hierinfo}) is already implied by the medical codes’ labels, the existing parent node is reused; otherwise, a new latent node is created. The node set is then updated to $\{ h_i \}_{1 \le i \le n_B}$, and a new distance matrix is recomputed using \eqref{addsuffequation}. The full recursive grouping procedure is summarized in Algorithm~\ref{treerecovery}.

\begin{algorithm}
\caption{Recursive Grouping with $k$-Means Clustering}
\label{treerecovery}
\begin{algorithmic}
    \STATE \textbf{Input:} Lorentzian distance matrix for node set $\mathcal{V}$
    \WHILE{stopping criterion not met}
        \STATE Compute $[D_{ij}]_{1 \le i,j \le |\mathcal{V}|}$ using \eqref{Dij} where $|\mathcal{V}|$ is the size of $\mathcal{V}$
        \STATE Cluster nodes into bottom sets $\{\mathcal{B}_i\}_{1 \le i \le n_B}$ using $k$-means clustering 
        \STATE Assign latent node $h_i$ for each $\mathcal{B}_i$; update node set $\mathcal{V} = \{h_i\}_{1 \le i \le n_B}$
        \STATE Recompute Lorentzian distance matrix using \eqref{addsuffequation}
    \ENDWHILE
\end{algorithmic}
\end{algorithm}

The hierarchical tree is automatically constructed by applying Algorithm~\ref{treerecovery} to the learned embeddings. Users may also personalize the high-level structure. For example, in the case of medications, the first layer of the tree could reflect drug class or therapeutic use. To support this, our framework allows user-defined categories at the top level, with the recursive algorithm applied within each category to generate a more detailed, data-driven hierarchy.

Finally, to improve interpretability, we use GPT-4o \citep{hurst2024gpt}, with prompt given in Supplementary Section~\ref{detailsetapp}, to annotate each latent node with a concise human-readable summary synthesized from its descendants. This enhances transparency and facilitates clinical validation of the discovered structure.

\section{Real-World Validation of MASH}
\label{sec:realdata}

We next evaluate MASH on the VA and MGB EHR datasets to assess both embedding quality and the accuracy of the constructed hierarchies. Our evaluation has two primary goals: (i) to demonstrate that MASH produces embeddings that integrate co-occurrence and semantic signals more effectively than existing approaches, and (ii) to show that the learned embeddings can be used to construct clinically coherent hierarchies, including codes that lack any standardized ontology. Across experiments, we compare MASH to a range of baseline and ablation methods and report results for diagnosis codes (PheCode), medication codes (RxNorm), and laboratory codes (LOINC and local codes).

\subsection{Evaluation of Embedding Quality and Utility}

\paragraph{\bf Baselines} We compare the quality of MASH embeddings against several baseline methods:  1) \textbf{CODER} \citep{yuan2022coder}: captures semantic information from text but does not incorporate co-occurrence patterns; 2) \textbf{MIKGI} \citep{zhou2022multiview}: integrates both co-occurrence statistics and textual data; 3)  \textbf{Procrustes alignment (PT)} \citep{kementchedjhieva-etal-2018-generalizing}: applies an optimal rotation matrix to align SPPMI-SVD embeddings across MGB and VA; and 4) \textbf{SDNE} \citep{wang2016structural}: structural deep network embedding applied to the aggregated SPPMI matrices from MGB and VA, as well as to the CODER correlation matrix (Supplementary Section~\ref{baselinesec}). For MASH embeddings, we include ablation studies by considering only the harmonized embeddings after OT as well as the subsequent hyperbolic embeddings.

\paragraph{\bf Evaluation Metrics}
To assess how well the embeddings capture clinically meaningful relationships, we compute the area under the ROC curve (AUC) for distinguishing known similar or related code pairs from random pairs based on embedding distance. For MASH, we use the Lorentzian inner product, while for other methods, we use cosine similarity. Evaluations are conducted both overall and separately within four code categories: PheCode–PheCode, RxNorm–RxNorm, Lab–Lab, and LOINC–local lab code pairs.  We additionally quantified the utility of the embeddings for selecting clinicallly relevant features for diseases of interest. To this end, we randomly selected 21 PheCodes to represent target diseases and identify the top $100$ features across different categories using embeddings from SPPMI-SVD (MGB and VA) and CODER - we select the features with the highest cosine similarity with the target disease. For each disease, we compute a relevance score between the disease and candidate features based on embedding distances for each embedding method. In parallel, GPT-4o is used to generate semantic relevance scores from feature descriptions (Supplementary~\ref{appmetrics}). Agreement is quantified using Spearman’s rank correlation between the embedding-based and GPT-4o–based rankings.  

\subsubsection{Embedding Quality Results}
Table~\ref{embeddingauctable} presents the AUC results. The optimal transport (OT) aggregation method consistently outperforms the baseline approaches, and the subsequent hyperbolic embedding step further boosts performance. Compared to CODER, the OT-based embeddings yield substantial gains in detecting clinically related pairs (e.g., functionally or ontologically linked codes), highlighting the benefit of incorporating co-occurrence information.  

\begin{table}[ht]
\centering
\resizebox{0.9\textwidth}{!}{%
\begin{tabular}{lcccccccccc}
\toprule
& \multicolumn{2}{c}{All codes} & \multicolumn{2}{c}{PheCode} & \multicolumn{2}{c}{RxNorm} & \multicolumn{1}{c}{Lab} & \multicolumn{1}{c}{LOINC to local lab code} \\
\cmidrule(r){2-3} \cmidrule(r){4-5} \cmidrule(r){6-7} \cmidrule(r){8-8} \cmidrule(r){9-9}
& sim & rel & sim & rel & sim & rel & sim &  \\
\midrule
SDNE & 0.857 & 0.859 & 0.877 & 0.869 & 0.800 & 0.774 & 0.846 & 0.789 \\
PT & 0.896 & 0.769 & 0.914 & 0.805 & 0.850 & 0.740 & 0.903 & 0.824 \\
MIKGI & 0.926 & 0.732 & 0.907 & 0.776 & 0.831 & 0.712 & 0.923 & 0.927 \\
CODER & 0.961 & 0.709 & 0.915 & 0.773 & 0.897 & 0.812 & 0.982 & 0.985 \\
MASH: OT Aggregated & 0.967 & 0.789 & 0.942 & 0.828 & 0.901 & 0.785 & 0.983 & 0.984 \\
MASH: Hyperbolic Embeddings & \textbf{0.982} & \textbf{0.889} & \textbf{0.962} & \textbf{0.886} & \textbf{0.958} & \textbf{0.885} & \textbf{0.992} & \textbf{0.986} \\
\bottomrule
\end{tabular}%
}
\caption{AUC for embedding quality across methods and code categories. ``MASH: OT Aggregated'' refers to raw Euclidean embeddings after optimal transport but before hyperbolic embedding; and "MASH: Hyperbolic Embeddings" represent the final MASH embeddings after further training into the hyperbolic space. ``sim'' = similarity-based labels; ``rel'' = relatedness-based labels.}
\label{embeddingauctable}
\end{table}

\subsubsection{Feature Selection Utility}

\begin{figure}[H]
    \centering
    \includegraphics[width=0.99\linewidth,angle=0]{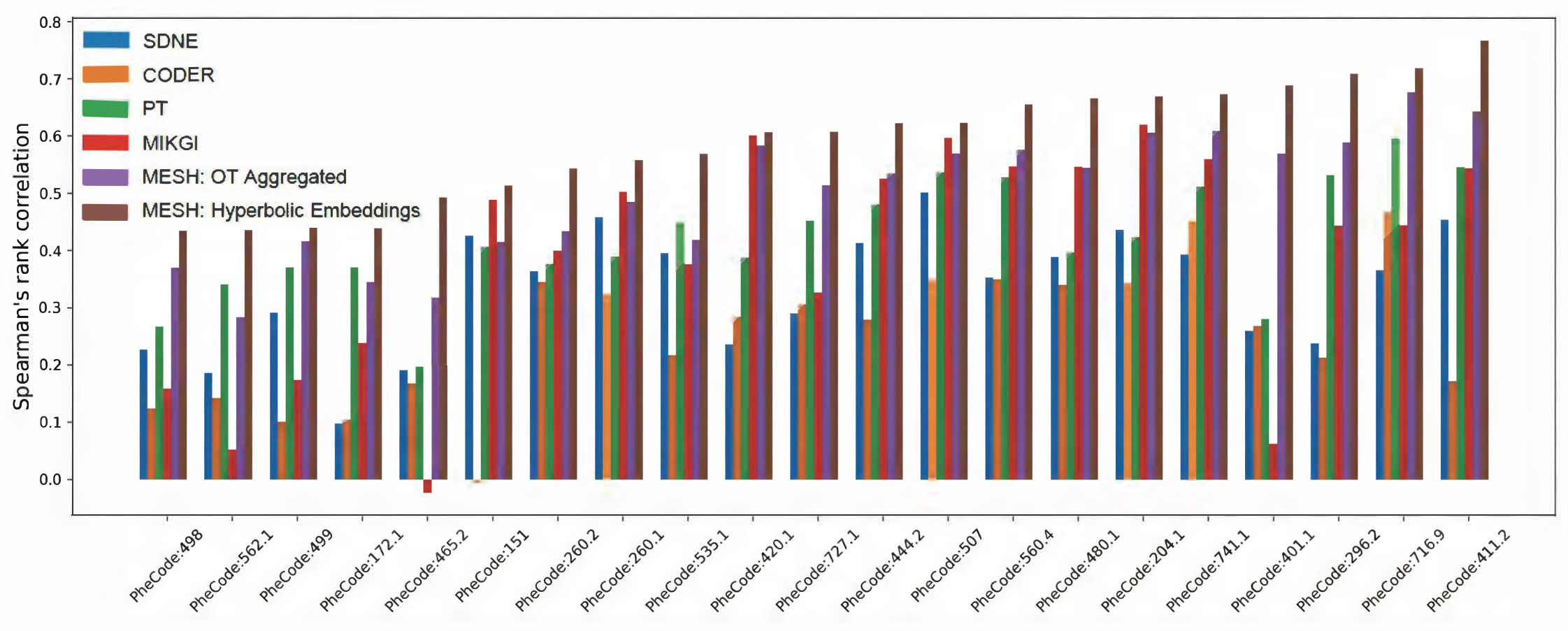}
    \caption{
    Feature selection performance. Average correlations: SDNE = $0.332$, CODER = $0.256$, PT = $0.421$, MIKGI = $0.390$, OT Aggregated = $0.500$, Hyperbolic Embeddings = \textbf{0.593}.}
    \label{rank_corr}
\end{figure}

Figure~\ref{rank_corr} shows the rank correlations between embedding-based feature importance scores and GPT-4o rankings across the selected PheCodes. The MASH embeddings, obtained through OT aggregation and further refined with hyperbolic training, achieve the strongest alignment with GPT-4o. The improvement over OT aggregation alone demonstrates the added value of hyperbolic refinement for enhancing feature selection capability.

\subsection{Evaluation of Constructed Hierarchy Graphs}

We evaluate the quality of hierarchy graphs constructed for diagnosis codes (PheCode), medication codes (RxNorm), and laboratory codes (LOINC and local). We compare to baseline methods by performing $k$-means clustering using the corresponding embeddings and then organized into a tree structure via hierarchical clustering. See Supplementary Section~\ref{baselinesec} for more details. 

\subsubsection{Evaluation Metrics}
We consider several evaluation metrics to assess the quality of the constructed hierarchies: 

\begin{itemize}
\item \textbf{Structural alignment with reference ontologies.}
For code types with established hierarchies (PheCode, RxNorm, and LOINC), we assess alignment with the reference ontology using \emph{Normalized Mutual Information (NMI)} and \emph{Adjusted Rand Index (ARI)}. These metrics quantify how well the partitioning induced by the constructed tree recovers the true class labels. Results are summarized in Table~\ref{nmi_ari_table}.

\item \textbf{Integration of local laboratory codes.}
Local laboratory codes do not have a predefined ontology. To evaluate their integration, we construct a reference tree by mapping $1897$ VA local laboratory codes to LOINC or LP concepts using a manually curated dictionary. As in the other settings, we report NMI and ARI scores to assess alignment with the curated structure.

\item \textbf{Fine-grained structural coherence.}
To further evaluate structural fidelity, we compute \emph{precision} and \emph{sensitivity} scores (Table~\ref{sensitivity_precision_table}). Precision is the probability that two codes classified as siblings in the constructed tree are also siblings in the reference tree. Sensitivity is the probability that true siblings in the reference tree are grouped together in the constructed tree. These metrics highlight the ability to capture fine-grained sibling relationships.

\item \textbf{Semantic interpretability.}
Beyond structure, we assess interpretability using GPT-4 annotations (Appendix~\ref{detailsetapp}). Each internal node in the constructed tree is labeled using our annotation module. Two scores are defined: 1)  \textbf{Hierarchy score}: For each internal node, GPT-4 assigns a score of 1 if all child nodes represent valid sub-concepts of the parent annotation and 0 otherwise. The hierarchy score is the average across all internal nodes; and 2) \textbf{Divergence score}: For each set of sibling nodes, GPT-4 assigns a score of 1 if all annotations are semantically distinct and 0 otherwise. The divergence score is the average across all sibling sets.  
\end{itemize}
\subsubsection{Results}

Across all code types and evaluation criteria, MASH consistently outperforms the baselines. As shown in Table~\ref{nmi_ari_table}, it achieves the highest NMI and ARI values, demonstrating strong alignment with standard ontologies. Table~\ref{sensitivity_precision_table} further shows that MASH produces hierarchies with substantially higher precision and sensitivity, reflecting accurate reconstruction of sibling relationships. For local lab codes, MASH attains the highest AUC, indicating effective integration of local codes with their mapped LOINC counterparts. Finally, the GPT-4–based semantic evaluation in Table~\ref{table_performance_comparison} reveals that MASH achieves near-perfect hierarchy and divergence scores, confirming that its constructed hierarchies are not only structurally faithful but also semantically coherent and interpretable.

\begin{table}[ht]
\centering
\resizebox{0.9\textwidth}{!}{%
\begin{tabular}{lcccccccccc}
\toprule
& \multicolumn{2}{c}{PheCode} & \multicolumn{2}{c}{RxNorm} & \multicolumn{2}{c}{Lab} & \multicolumn{2}{c}{LOINC} & \multicolumn{2}{c}{Local lab} \\
\cmidrule(r){2-3} \cmidrule(r){4-5} \cmidrule(r){6-7} \cmidrule(r){8-9} \cmidrule(r){10-11} 
& NMI & ARI & NMI & ARI & NMI & ARI & NMI & ARI & NMI & ARI  \\
\midrule
SDNE & 0.908 & 0.431 & 0.811 & 0.365 & 0.880 & 0.336 & 0.894 & 0.122 & 0.863 & 0.521  \\
PT & 0.917 & 0.394 & 0.792 & 0.225 & 0.881 & 0.271 & 0.907 & 0.420 & 0.821 & 0.152  \\
MIKGI & 0.905 & 0.300 & 0.758 & 0.285 & 0.878 & 0.214 & 0.907 & 0.317 & 0.815 & 0.146  \\
CODER & 0.901 & 0.278 & 0.808 & 0.342 & 0.909 & 0.307 & 0.926 & 0.359 & 0.879 & 0.341  \\
\textbf{MSH} & \textbf{0.942} & \textbf{0.606} & \textbf{0.927} & \textbf{0.719} & \textbf{0.930} & \textbf{0.569} & \textbf{0.947} & \textbf{0.719} & \textbf{0.892} & \textbf{0.638}  \\
\bottomrule
\end{tabular}%
}
\caption{Comparison of NMI and ARI across methods for different code types. ``Lab'' denotes the integration of LOINC and local laboratory codes. }
\label{nmi_ari_table}
\end{table}

\begin{table}[ht]
\centering
\scalebox{0.9}{%
\begin{tabular}{lcccccccccc}
\toprule
& \multicolumn{2}{c}{PheCode} & \multicolumn{2}{c}{RxNorm} & \multicolumn{2}{c}{Lab} & \multicolumn{2}{c}{LOINC} & \multicolumn{2}{c}{Local lab} \\
\cmidrule(r){2-3} \cmidrule(r){4-5} \cmidrule(r){6-7} \cmidrule(r){8-9} \cmidrule(r){10-11}
& sen & pre & sen & pre & sen & pre & sen & pre & sen & pre \\
\midrule
SDNE & 0.538 & 0.362 & 0.498 & 0.256 & 0.478 & 0.305 & 0.192 & 0.103 & 0.252 & 0.853 \\
PT & 0.327 & 0.499 & 0.396 & 0.177 & 0.210 & 0.455 & 0.460 & 0.413 & 0.119 & 0.626 \\
MIKGI & 0.238 & 0.409 & 0.478 & 0.187 & 0.183 & 0.411 & 0.322 & 0.329 & 0.148 & 0.650 \\
CODER & 0.214 & 0.399 & 0.419 & 0.261 & 0.229 & 0.605 & 0.336 & 0.447 & 0.241 & 0.861 \\
\textbf{MASH} & \textbf{0.776} & \textbf{0.498} & \textbf{0.720} & \textbf{0.595} & \textbf{0.536} & \textbf{0.724} & \textbf{0.794} & \textbf{0.662} & \textbf{0.440} & \textbf{0.867} \\
\bottomrule
\end{tabular}%
}
\caption{Comparison of sensitivity (sen) and precision (pre) across methods for different code types. ``Lab'' denotes the integration of LOINC and local laboratory codes.}
\label{sensitivity_precision_table}
\end{table}

\begin{table}[ht]
\centering
\scalebox{0.9}{%
\begin{tabular}{lcc}
\toprule
         & Hierarchy & Divergence  \\ 
\midrule
PheCode  & 0.98      & 1          \\ 
RxNorm   & 0.99      & 1           \\ 
Lab      & 0.97      & 0.99       \\ 
\bottomrule
\end{tabular}%
}
\caption{Comparison of semantic annotation quality using GPT-4. ``Hierarchy'' measures consistency between parent and child annotations; ``Divergence'' quantifies distinctiveness among sibling annotations.}
\label{table_performance_comparison}
\end{table}

\subsubsection{Case Studies of Constructed Hierarchy Graphs}
\label{casestudy}

We present qualitative case studies of hierarchy graphs constructed by MASH, with interactive visualizations of the full hierarchies available at \url{https://celehs.github.io/MASH/}.  

\begin{figure}[H]
\begin{center}
    \subfloat[PheCode hierarchy]{\includegraphics[width=0.5\textwidth]{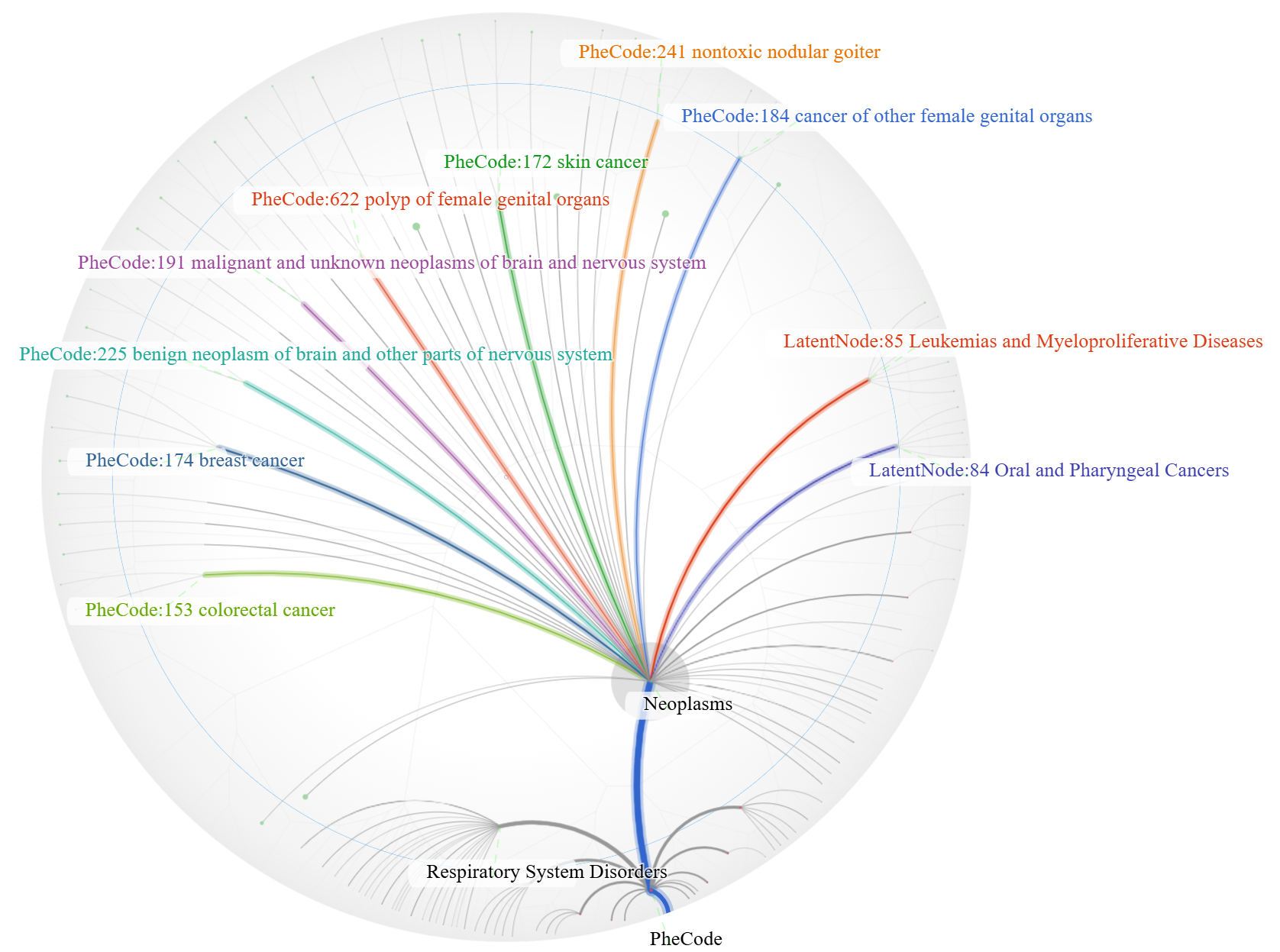}\label{cphecode}} 
    \hfill
    \subfloat[RxNorm hierarchy]{\includegraphics[width=0.5\textwidth]{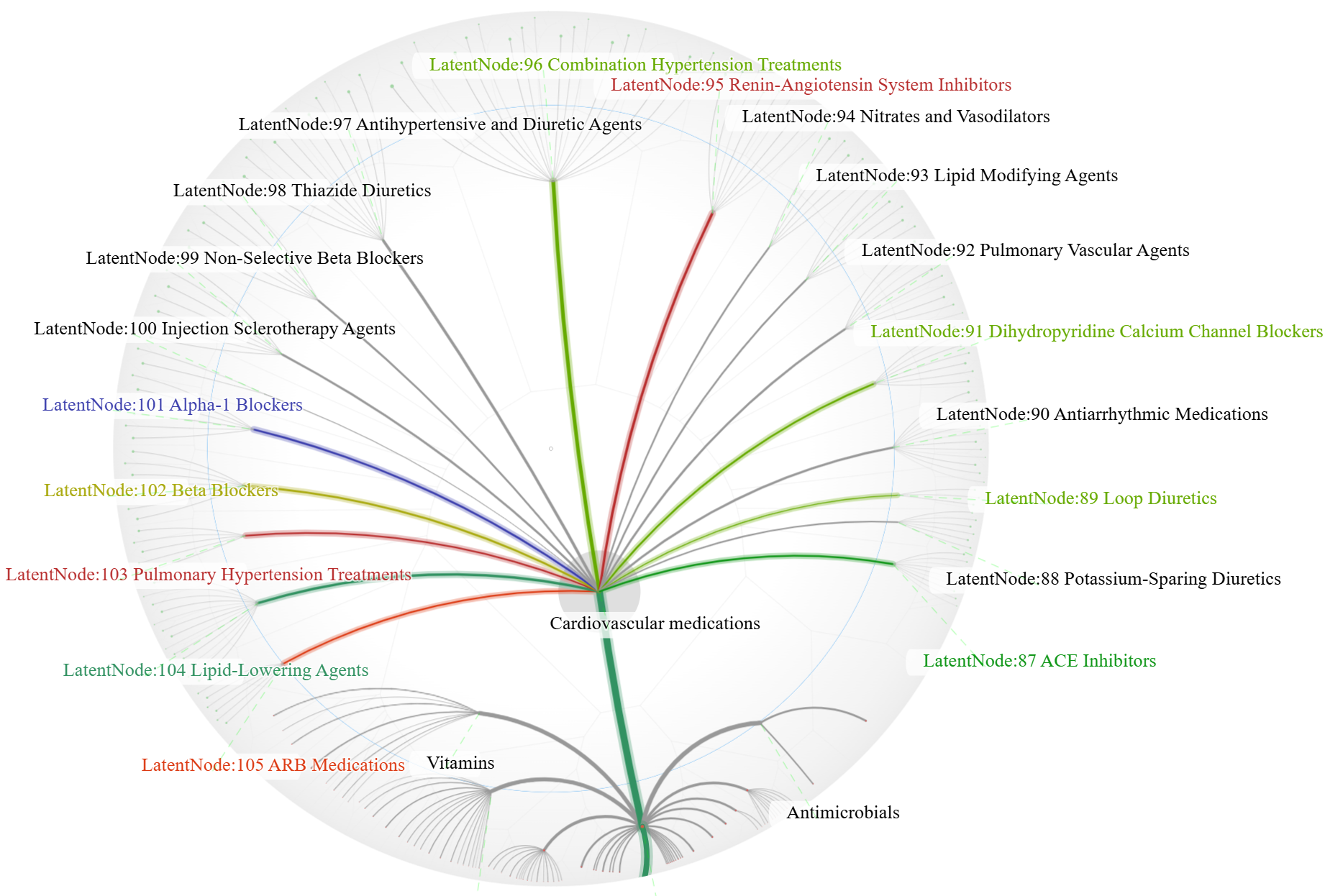}\label{crxnorm}} 
    \hfill
    \subfloat[Lab hierarchy]{\includegraphics[width=0.5\textwidth]{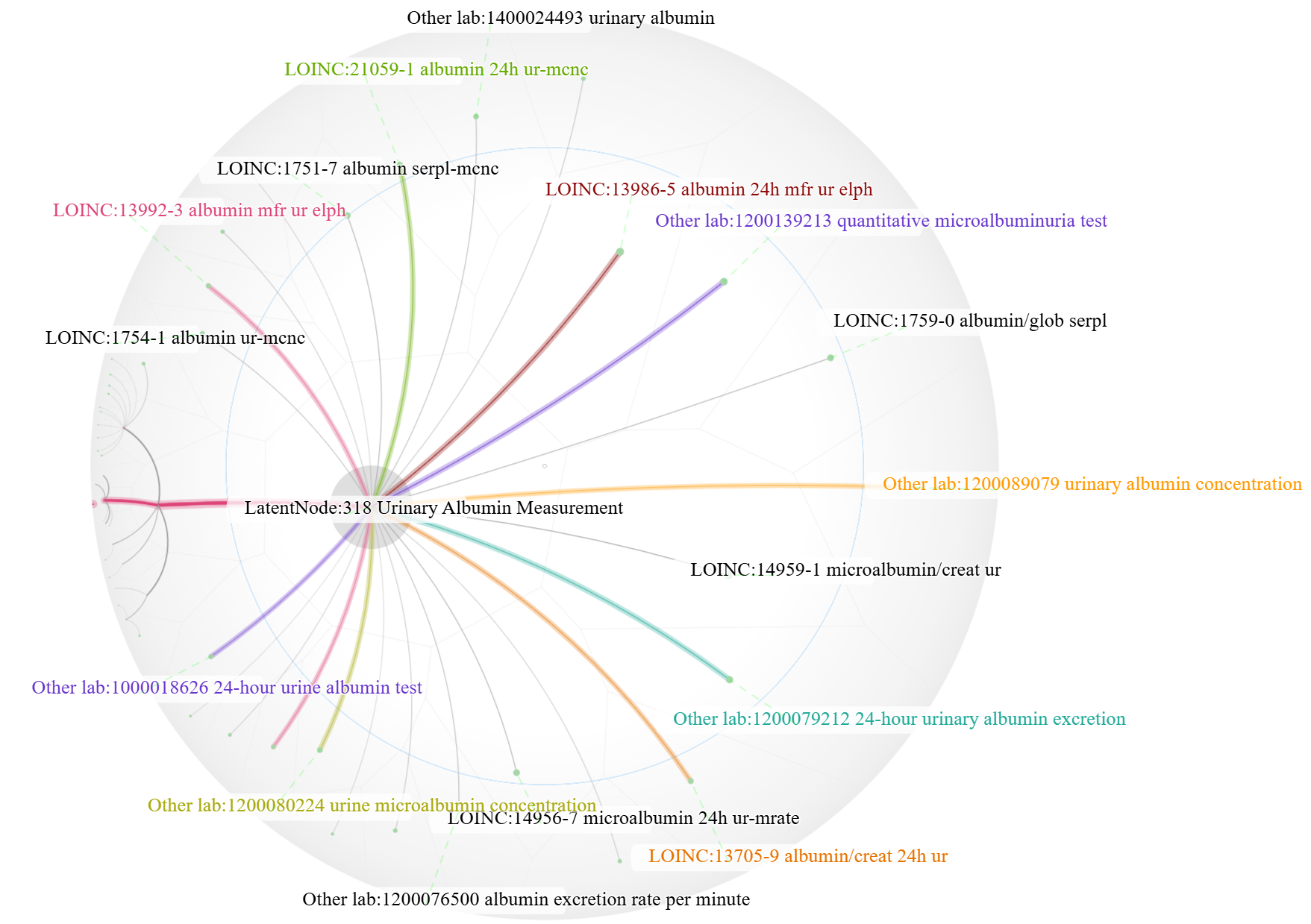}\label{clab}}
\end{center}    
    \caption{Partial views of constructed hierarchy graphs for (a) diagnosis codes (PheCode), (b) medication codes (RxNorm), and (c) laboratory test codes (LOINC and local lab codes).}
    \label{hierarchy_case}
\end{figure}

Figure~\ref{hierarchy_case} shows partial views of the hierarchies for diagnoses (PheCodes), medications (RxNorm), and laboratory tests (LOINC and local codes). In the PheCode graph, major disease categories such as breast cancer (PheCode 174), colorectal cancer (153), and skin cancer (172) are clearly separated, while subtypes such as benign and malignant brain neoplasms (225, 191) and clusters of leukemias and oral/pharyngeal cancers emerge as coherent subtrees. The RxNorm graph recovers pharmacological classes such as Beta Blockers, ACE Inhibitors, Loop Diuretics, and Calcium Channel Blockers, with finer categories (e.g., Thiazide Diuretics, Non-Selective Beta Blockers) nested appropriately. In the laboratory test graph, nodes such as Urinary Albumin Measurement correctly align semantically related LOINC codes (e.g., 13992-3, 13986-5, 21059-1) with local codes (e.g., 1200139213, 1200079212). These case studies highlight MASH’s ability to recover clinically meaningful structures across heterogeneous code systems, including integration of standardized and local codes without explicit supervision.  

\section{Discussion}
\label{sec:discussion}

In this work, we introduced MASH, a unified framework for aligning multi-source medical information and constructing hierarchical knowledge graphs. A central contribution of MASH is its neural network–based optimal transport framework, which explicitly aligns embeddings across institutions while preserving both co-occurrence patterns and semantic structure. This principled approach to handling between-source heterogeneity makes MASH particularly well-suited for cross-institutional applications, where variation in coding practices and patient populations has often limited the generalizability of prior methods. By integrating heterogeneous data streams into a shared embedding space, MASH provides a robust and scalable foundation for downstream clinical and translational research.  

Another key innovation lies in transforming these embeddings into hyperbolic space through a semi-supervised learning framework. By leveraging existing ontologies where available, and inferring structure where none exists, MASH can respect curated knowledge while automatically extending hierarchies to incorporate local and unmapped codes. This semi-supervised design has important implications for health systems, where standardized vocabularies often fail to capture the breadth of local practice or keep pace with evolving medical knowledge. In this way, MASH supports a dynamic and continuously updating hierarchy that aligns with the vision of a learning health system rather than a static ontology.  

Our evaluation demonstrates that MASH consistently recovers established structures across diagnoses, medications, and laboratories, while also producing clinically coherent graphs in domains without gold-standard hierarchies. The ability to integrate local codes, enrich them with semantic context, and situate them within a broader hierarchical framework has direct relevance for interoperability, cohort discovery, and the development of more interpretable AI models. Furthermore, the integration of pre-trained language models for annotation highlights the potential of combining foundation models with structured EHR data to generate clinically meaningful and human-readable knowledge artifacts. Taken together, these findings suggest that MASH is a step toward adaptive, data-driven ontologies that can evolve alongside both clinical practice and biomedical discovery.  

Looking forward, MASH could be further refined by more effectively leveraging advances in large language models such as GPT-5. These models are increasingly capable of providing partial label information, resolving ambiguous mappings, and generating semantically coherent annotations at scale. Incorporating such partial supervision into the hyperbolic learning framework could improve hierarchy construction in areas where curated ontologies are incomplete or missing. Moreover, large language models can support automated evaluation as part of the knowledge distillation process, offering a scalable mechanism to continuously assess the semantic coherence of newly integrated codes. By embedding these capabilities into MASH, the framework could evolve into a more adaptive system that not only aligns heterogeneous EHR data and local codes, but also integrates real-time feedback from language models to maintain clinically meaningful and dynamically updated hierarchies.

\section*{Data Availability Statement}
We validate the efficacy of our algorithm using electronic health record (EHR) data from the U.S. Department of Veterans Affairs (VA) and Mass General Brigham (MGB) Healthcare Systems. These data are protected under federal privacy regulations and cannot be shared publicly. The authors do not have authority to distribute the underlying VA and MGB EHR data.

\bibliographystyle{chicago}
\bibliography{reference}

\clearpage
\appendix

\renewcommand{\thetable}{S\arabic{table}}
\renewcommand{\thefigure}{S\arabic{figure}}
\renewcommand{\theequation}{S\arabic{equation}}
\setcounter{equation}{0}
\setcounter{table}{0}
\setcounter{figure}{0}

\begin{center}
	{\bf  \Large
		Supplementary Material for ``Automated Hierarchical Graph Construction for Multi-source Electronic Health Records''}  \\
\end{center}

The supplementary material is organized as follows. 
Section~\ref{expdetail} provides experimental details. Section~\ref{thecla} contains detailed discussion about the additivity loss in Sections \ref{subsehyplearn} and criterion for hierarchy construction in Section \ref{subsehiercluster}.

\section{Experimental Details}
\label{expdetail}

\subsection{MASH Settings} 
\label{detailsetapp}

\paragraph{Hyperbolic Learning} We set the weighting parameters in the loss function as $w_a = 0.1$, $w_c = 0.1$, and $w_e = 1$, where the additivity loss is computed for PheCode, RxNorm, and all laboratory codes. The learning rate is set to be $0.01$. We train for a total of $1000$ epochs. During the first $100$ warm-up epochs, we only fine-tune the embeddings of PheCode and RxNorm. The initial embeddings are the $d = 300$ aggregated embeddings obtained using optimal transport.

\paragraph{Graph Construction} In each iteration of Algorithm~\ref{treerecovery}, after deriving the distance matrix $D$, we perform $k$-means clustering, with $k$ chosen according to the Silhouette score. To make the graph more interpretable and the clustering more efficient, we first define the general categories, then construct the graph within each category. We define 21 general categories for PheCode (Table~\ref{phecodecat}) based on disease types, and 32 general categories for RxNorm (Table~\ref{rxnormcat}) based on the drug categories in the VA system. The LOINC hierarchy in the LOINC Multiaxial Hierarchy framework can reach a depth of 13, which contains overly redundant binary classifications. Therefore, we use only the first 5 levels for our pre-classification.

\begin{table}[h!]
\centering
\caption{Number of PheCodes in Each General Disease Category}
\begin{tabular}{llr}
\toprule
\textbf{Index} & \textbf{General Category} & \textbf{Number of Codes} \\
\midrule
0  & Circulatory System Disorders                     & 157 \\
1  & Complications of Medical and Surgical Care       & 21  \\
2  & Congenital and Genetic Disorders                 & 54  \\
3  & Digestive System Disorders                       & 125 \\
4  & Endocrine, Nutritional, and Metabolic Disorders  & 120 \\
5  & Genitourinary System Disorders                   & 94  \\
6  & Hematologic Disorders                            & 61  \\
7  & Immune System Disorders                          & 9   \\
8  & Infectious Diseases                              & 67  \\
9  & Injury and Poisoning                             & 100 \\
10 & Maternal and Perinatal Complications             & 46  \\
11 & Mental Health Disorders                          & 53  \\
12 & Musculoskeletal and Connective Tissue Disorders  & 160 \\
13 & Neoplasms                                        & 157 \\
14 & Neurological Disorders                           & 90  \\
15 & Oral and Maxillofacial Disorders                 & 48  \\
16 & Others and Ambiguous                             & 146 \\
17 & Otolaryngology and Ophthalmology Disorders       & 143 \\
18 & Reproductive System Disorders                    & 61  \\
19 & Respiratory System Disorders                     & 68  \\
20 & Skin and Subcutaneous Tissue Disorders           & 75  \\
\bottomrule
\end{tabular}
\label{phecodecat} 
\end{table}

\begin{longtable}{llr}
\caption{Number of RxNorm in Each General Category} \label{rxnormcat} \\
\toprule
\textbf{Index} & \textbf{Level 2 Category} & \textbf{Number of Codes} \\
\midrule
\endfirsthead

\multicolumn{3}{c}%
{{\bfseries \tablename\ \thetable{} -- continued from previous page}} \\
\toprule
\textbf{Index} & \textbf{Level 2 Category} & \textbf{Number of Codes} \\
\midrule
\endhead

\midrule \multicolumn{3}{r}{{Continued on next page}} \\
\endfoot

\bottomrule
\endlastfoot

0  & Antidotes, deterrents and poison control            & 16  \\
1  & Antihistamines                                      & 14  \\
2  & Antimicrobials                                      & 140 \\
3  & Antineoplastics                                     & 107 \\
4  & Antiparasitics                                      & 20  \\
5  & Antiseptics/disinfectants                           & 1   \\
6  & Autonomic medications                               & 30  \\
7  & Blood products/modifiers/volume expanders           & 42  \\
8  & Cardiovascular medications                          & 152 \\
9  & Central nervous system medications                  & 200 \\
10 & Dental and oral agents, topical                     & 13  \\
11 & Dermatological agents                               & 168 \\
12 & Diagnostic agents                                   & 22  \\
13 & Gastrointestinal medications                        & 98  \\
14 & Genitourinary medications                           & 27  \\
15 & Herbs/alternative therapies                         & 27  \\
16 & Hormones/synthetics/modifiers                       & 124 \\
17 & Immunological agents                                & 49  \\
18 & Intrapleural medications                            & 1   \\
19 & Irrigation/dialysis solutions                       & 5   \\
20 & Miscellaneous agents                                & 2   \\
21 & Musculoskeletal medications                         & 61  \\
22 & Nasal and throat agents, topical                    & 29  \\
23 & Ophthalmic agents                                   & 104 \\
24 & Otic agents                                         & 13  \\
25 & Pharmaceutical aids/reagents                        & 15  \\
26 & Prosthetics/supplies/devices                        & 21  \\
27 & Rectal, local                                       & 16  \\
28 & Respiratory tract medications                       & 82  \\
29 & Therapeutic nutrients/minerals/electrolytes         & 67  \\
30 & Unknown                                             & 449 \\
31 & Vitamins                                            & 37  \\
\end{longtable}

\paragraph{GPT Annotation} After obtaining the hierarchy graph, we use GPT-4o to annotate its latent nodes. Note that all leaf nodes (medical codes) in the hierarchy already have predefined descriptions. Specifically, for each identified three-layer structure, we prompt GPT to annotate the middle-layer latent nodes—ensuring they share a common theme yet remain distinct, since they represent different subcategories. After annotating the middle layer, we delete the leaf nodes in that three-layer structure from the graph and then repeat this process until no three-layer structures remain. The exact prompts are shown in Table~\ref{tab:annotation-prompts}.

\begin{table}[H]
\centering
\caption{Annotation prompts used for labeling latent nodes in the hierarchy.}
\label{tab:annotation-prompts}
\setlength{\tabcolsep}{6pt}
\renewcommand{\arraystretch}{1.2}
\begin{tabular}{@{}p{0.16\linewidth} p{0.78\linewidth}@{}}
\toprule
\textbf{Domain} & \textbf{Prompt} \\
\midrule
PheCode &
\texttt{\{data\}} \par\vspace{0.4ex}
Each cluster contains codes paired with descriptions in the format: \texttt{`code:description'}. Some clusters lack descriptions. Use your knowledge of Phecodes and language comprehension skills to give a summary phrase (few words) for each cluster. These cluster summary phrases should serve as classifications for the clusters, each distinctly different from the others. Please respond with \texttt{`cluster:description'} for each cluster, separated by \texttt{\textbackslash n}, with no other words or characters. \\
\addlinespace[0.6ex]
RxNorm &
\texttt{\{data\}} \par\vspace{0.4ex}
Each cluster contains codes paired with descriptions in the format: \texttt{`code:description'}. Some clusters lack descriptions. Use your knowledge of RxNorm codes and language comprehension skills to give a summary phrase (few words) for each cluster. These cluster summary phrases should serve as classifications for the clusters, each distinctly different from the others. Please respond with \texttt{`cluster:description'} for each cluster, separated by \texttt{\textbackslash n}, with no other words or characters. \\
\addlinespace[0.6ex]
Lab Codes &
\texttt{\{data\}} \par\vspace{0.4ex}
Each cluster contains codes paired with descriptions in the format: \texttt{`code:description'}. Some clusters lack descriptions. Use your language comprehension skills to give a summary phrase (few words) for each cluster. These cluster summary phrases should serve as classifications for the clusters, each distinctly different from the others. Please respond with \texttt{`cluster:description'} for each cluster, separated by \texttt{\textbackslash n}, with no other words or characters. \\
\bottomrule
\end{tabular}
\end{table}

\subsection{Baseline Settings}
\label{baselinesec}

\paragraph{MIKGI} We follow MIKGI's approach to aggregate the CODER embedding, VA and MBG embeddings ($d = 300$).

\paragraph{Procrustes Analysis} Given the $d = 300$ embeddings of CODER, VA, and MGB, we first compute the transformation matrices between CODER and VA, and between CODER and MGB. We then obtain the transformed embeddings of VA and MGB. Finally, we aggregate the transformed VA and MGB embeddings with the CODER embeddings by averaging them, followed by normalization.

\paragraph{SDNE} We first compute the correlation matrix of the CODER embeddings ($d = 300$) and set all elements less than or equal to the 90th percentile to zero. Similarly, we zero out the elements less than or equal to the 90 percentile in the SPPMI matrices of VA and MGB. We then add the three matrices together and again set elements less than or equal to the 90th percentile to zero. Finally, we apply a neural network with hidden layers of sizes 2000, 1000, 500, 500, and 300 to derive the $d = 300$ embeddings.

\paragraph{Hierarchical Clustering}
We first apply classic $k$-means clustering to group the codes, followed by hierarchical clustering on the resulting clusters to generate the final tree structure.
For the hierarchical clustering step, we use the \texttt{scipy.cluster.hierarchy.linkage} function with the \texttt{average} method, which corresponds to unweighted pair group method with arithmetic mean (UPGMA). The input to \texttt{linkage} is the condensed pairwise distance matrix computed from the embeddings.

\subsection{Metrics}
\label{appmetrics}

\paragraph{Feature Selection} We select 21 PheCodes, including PheCodes 172.1, 204.1, 151, 420.1, 401.1, 444.2, 411.2, 480.1, 498, 465.2, 499, 507, 535.1, 560.4, 562.1, 260.1, 296.2, 260.2, 716.9, 741.1, and 727.1. The correlation is reported using Spearman rank correlation. We use the following prompt to provide the scores. 

\begin{tcolorbox}[
  width       = \linewidth,        
  enhanced, breakable,
  colback     = blue!5!white,
  colframe    = gray,
  title       = Prompt format example,
  fonttitle   = \bfseries\footnotesize,
  sharp corners,
  parbox      = false,
  before upper={%
      \setlength{\parindent}{0pt}
      \small\ttfamily\raggedright  
  }
]
Is the following medical code/description related to 
\{\textit{target\_code\_desc}\}?\\[1ex]

\{\textit{medical\_code\_desc}\}\\[1ex]

Please provide a score between 0 and 1 indicating the likelihood that this code is related to \{\textit{target\_code\_desc}\}. For instance, a score of~1 indicates complete relevance, whereas a score of~0 indicates no relevance. Provide your score as a decimal (e.g.\ 0.13, 0.75, etc.) without any additional information or explanations.
\end{tcolorbox}

\paragraph{Hierarchy Score and Divergence Score}

We derive the Hierarchy Score and Divergence Score using the following prompts.

\begin{tcolorbox}[
  width       = \linewidth,
  enhanced, breakable,
  colback     = blue!5!white,
  colframe    = gray,
  title       = Prompt for Hierarchy Score,
  fonttitle   = \bfseries\footnotesize,
  sharp corners,
  parbox      = false,
  before upper={%
      \setlength{\parindent}{0pt}%
      \small\ttfamily\raggedright
  }
]
I have a main conception ``\{main\_conception\}'' and \{num\_subs\} sub-conceptions: \{sub\_1\}, \{sub\_2\}, \{sub\_3\}, \dots.  
Tell me how many sub-conceptions are not a sub-conception of the main conception.  
Answer with the digit number only, with no other characters needed.
\end{tcolorbox}

\begin{tcolorbox}[
  width       = \linewidth,
  enhanced, breakable,
  colback     = blue!5!white,
  colframe    = gray,
  title       = Prompt for Divergence Score,
  fonttitle   = \bfseries\footnotesize,
  sharp corners,
  parbox      = false,
  before upper={%
      \setlength{\parindent}{0pt}%
      \small\ttfamily\raggedright
  }
]
I have \texttt{\{num\_conceptions\}} conceptions: \texttt{\{conception\_1\}}, \texttt{\{conception\_2\}}, \dots, \texttt{\{conception\_n\}}.\\[1ex]
Tell me if they are all semantically different (even subtle difference). Answer with \texttt{"Yes"} or \texttt{"No"}, with no other characters needed.
\end{tcolorbox}

\subsection{Simulations}
\label{sec:simulation}

To evaluate the robustness of our method and facilitate replication of our results (since the VA and MGB SPPMI matrix data are not publicly available), we simulate SPPMI matrices for PheCode and replicate the entire pipeline, including all baseline methods, to obtain comparable results.

Specifically, we first use the original $d = 300$ CODER embeddings to derive the correlation matrix $M$. From $M$, we extract submatrices corresponding to RPDR, VA, and the union of all codes. Each submatrix is then modified using a label-guided update strategy based on given positive and negative code pairs.
Let $(i,j)$ be a pair of codes annotated as either positively or negatively correlated. We perform the following updates:
\[
\begin{aligned}
&\text{If } (i,j) \text{ is a negative pair:} && M_{ij} \leftarrow \max(0, M_{ij} - \delta^-), \quad M_{ji} \leftarrow \max(0, M_{ji} - \delta^-), \\
&\text{If } (i,j) \text{ is a positive pair:} && M_{ij} \leftarrow \min(1, M_{ij} + \delta^+), \quad M_{ji} \leftarrow \min(1, M_{ji} + \delta^+).
\end{aligned}
\]

To systematically modify the similarity matrices, we introduce the following hyperparameters:

\begin{itemize}
    \item \texttt{neg\_frac}: The fraction of negatively labeled code pairs randomly selected from the positive pairs for similarity reduction.
    \item \texttt{pos\_frac}: The fraction of positively labeled code pairs randomly selected from the negative pairs (non-positive) for similarity enhancement.
    \item \texttt{neg\_shift}: The magnitude by which the similarity score is reduced for selected negative pairs.
    \item \texttt{pos\_shift}: The magnitude by which the similarity score is increased for selected positive pairs.
    \item \texttt{random seed}: The random seed used to ensure reproducibility of the pair-sampling process.
\end{itemize}

Specifically, we constructed three correlation matrices with distinct parameter settings:

\begin{itemize}
    \item \textbf{MGB}: $\texttt{neg\_frac}=0.1$, $\texttt{pos\_frac}=0.2$, $\texttt{neg\_shift}=0.1$, $\texttt{pos\_shift}=0.2$, and a random seed of $10$.
    \item \textbf{VA}: The same parameters as the RPDR matrix, except with a random seed of $20$.
    \item \textbf{CODER}: $\texttt{neg\_frac}=0.2$, $\texttt{pos\_frac}=0.4$, $\texttt{neg\_shift}=0.2$, $\texttt{pos\_shift}=0.3$, and a random seed of $30$.
\end{itemize}

Then we follow our pipelines and baselines as in the real world data analysis, to obtain the consistant results on the quality of embeddings (Table~\ref{sim:phecode_hierarchy}) and constructed graph (Table~\ref{sim:phecode_clustering}).

\begin{table}[h]
\centering
\begin{tabular}{lcc}
\toprule
\textbf{Method} & \textbf{sim} & \textbf{rel} \\
\midrule
SDNE                 & 0.757 & 0.814 \\
PT                   & 0.790 & 0.614 \\
MIKGI                & 0.738 & 0.495 \\
CODER                & 0.837 & 0.637 \\
OT Aggregated         & 0.889 & 0.732 \\
Hyperbolic Embedding & 0.957 & 0.872 \\
\bottomrule
\end{tabular}
\caption{Simulation results for derived PheCode embeddings using various methods.}
\label{sim:phecode_hierarchy}
\end{table}

\vspace{1em}

\begin{table}[h]
\centering
\begin{tabular}{lcccc}
\toprule
\textbf{Method} & \textbf{NMI} & \textbf{ARI} & \textbf{sen} & \textbf{pre} \\
\midrule
SDNE   & 0.8628 & 0.1886 & 0.3218 & 0.1359 \\
PT     & 0.9077 & 0.3501 & 0.2928 & 0.4388 \\
MIKGI  & 0.8971 & 0.3121 & 0.2821 & 0.3525 \\
CODER  & 0.9006 & 0.3229 & 0.2747 & 0.3951 \\
Ours   & 0.9392 & 0.5702 & 0.8070 & 0.4423 \\
\bottomrule
\end{tabular}
\caption{Simulation results for constructed PheCode graphs using various methods.}
\label{sim:phecode_clustering}
\end{table}

\section{Discussions on Latent Trees and Hierarchy Recovery} 
\label{thecla}

\subsection{Latent Tree Models and Information Distance}

A \textit{latent tree} \citep{cowell1999probabilistic} is defined as a tree with node set $W := V \cup H$, where $V$ denotes the set of observed nodes (with cardinality $m = |V|$), and $H$ represents the set of latent (hidden) nodes. Tree-structured graphical models form an essential and tractable class within probabilistic graphical models. Specifically, a multivariate probability distribution is Markov on an undirected tree $T_p = (W,E_p)$ if it satisfies the factorization property
\[
p(x_1, \dots, x_M) = \prod_{i \in W} p(x_i) \prod_{(i,j) \in E} \frac{p(x_i,x_j)}{p(x_i)p(x_j)},
\]
so that marginal probabilities $p(x_i)$ and pairwise joint probabilities $p(x_i,x_j)$ fully characterize the joint distribution.

We use the concept of \textit{information distance} to measure similarity between random variables. For two discrete variables $X_i$ and $X_j$, define the joint probability matrix
\begin{equation}
J_{ij}^{ab} = p(x_i = a, x_j = b), \quad a,b \in \mathcal{X},
\end{equation}
and the diagonal marginal probability matrix for variable $X_i$ as:
\begin{equation}
M_i^{aa} = p(x_i = a), \quad a \in \mathcal{X}.
\end{equation}
The \textit{information distance} $d_{ij}$ between variables $X_i$ and $X_j$ is then defined by:
\begin{equation}
d_{ij} = -\log \frac{|\det J_{ij}|}{\sqrt{\det M_i \det M_j}}.
\end{equation}

A key property is that information distances are \textbf{additive} along tree paths \citep{erdos1999few}: for any pair of nodes $k,l \in W$,
\[
d_{kl} = \sum_{(i,j) \in \text{Path}(k,l)} d_{ij}.
\]
That is, the distance between two nodes equals the sum of edge distances along their unique path in the tree. This principle underlies our hierarchy recovery approach.

\subsection{Additivity Condition and Its Implications}
\label{additivity condition}

We next elaborate why the sufficient additivity condition is enough to ensure the general path-based additivity. Specifically, assume that whenever $\mathbf{z}_i$ is the parent of $\mathbf{z}_j$ and $\mathbf{z}_k$ is not a descendant of $\mathbf{z}_j$, the relation
\[
d_{\ell}(\mathbf{z}_j, \mathbf{z}_k) = d_{\ell}(\mathbf{z}_i, \mathbf{z}_k) + d_{\ell}(\mathbf{z}_i, \mathbf{z}_j)
\]
holds. Under this condition, one can show that the more general formula
\[
d_{\ell}(\mathbf{z}_k, \mathbf{z}_l) = \sum_{(u,v) \in \text{Path}(k,l)} d_{\ell}(\mathbf{z}_u, \mathbf{z}_v)
\]
is satisfied for all node pairs $(k,l)$.

The argument proceeds by induction on the number of nodes in the tree.

\textbf{Base case:} Consider a tree with three nodes arranged in a chain (e.g., $\mathbf{z}_k$—$\mathbf{z}_i$—$\mathbf{z}_j$). If $\mathbf{z}_i$ is the parent of both $\mathbf{z}_j$ and $\mathbf{z}_k$, then the condition implies
\[
d_{\ell}(\mathbf{z}_j, \mathbf{z}_k) = d_{\ell}(\mathbf{z}_i, \mathbf{z}_k) + d_{\ell}(\mathbf{z}_i, \mathbf{z}_j),
\]
which exactly matches the additivity rule along the path $(\mathbf{z}_j \rightarrow \mathbf{z}_i \rightarrow \mathbf{z}_k)$.

\textbf{Inductive step:} Suppose the additivity property holds for all trees with up to $n$ nodes that satisfy the sufficient condition. Now consider a tree with $n+1$ nodes. Since every tree is connected and acyclic, there exists at least one leaf node. Remove a leaf $\mathbf{z}_j$ with parent $\mathbf{z}_i$, and let $\mathbf{z}_k$ be any node not in the subtree rooted at $\mathbf{z}_j$. By the sufficient condition we have
\[
d_{\ell}(\mathbf{z}_j, \mathbf{z}_k) = d_{\ell}(\mathbf{z}_i, \mathbf{z}_j) + d_{\ell}(\mathbf{z}_i, \mathbf{z}_k).
\]
By the inductive hypothesis, the distance $d_{\ell}(\mathbf{z}_i, \mathbf{z}_k)$ is additive along the path between $\mathbf{z}_i$ and $\mathbf{z}_k$ in the reduced tree with $n$ nodes. Hence, for any node $l$,
\[
d_{\ell}(\mathbf{z}_j, \mathbf{z}_l) 
= d_{\ell}(\mathbf{z}_j, \mathbf{z}_i) 
+ \sum_{(u,v) \in \text{Path}(i,l)} d_{\ell}(\mathbf{z}_u, \mathbf{z}_v).
\]
This shows that $d_{\ell}(\mathbf{z}_j, \mathbf{z}_l)$ is additive along the extended path including the edge $(\mathbf{z}_i, \mathbf{z}_j)$. By induction, the additivity condition holds for all pairs $(k,l) \in W$.

\subsection{Further Discussion on the Criterion $D_{ij}$}
\label{subsec:Dij}

The distance $D_{ij}$ in \eqref{Dij} plays a central role in identifying bottom sets during recursive grouping. Here we provide more details to justify why $D_{ij}=0$ is equivalent to nodes $i$ and $j$ belonging to the same bottom set, under the additivity property described in Section~\ref{additivity condition}.

\paragraph{Forward direction (same bottom set $\Rightarrow D_{ij}=0$).}
Suppose nodes $i$ and $j$ are in the same bottom set. By definition, they are either siblings or in a parent–child relationship. For any other node $k \in \mathcal{V}\setminus\{i,j\}$, we consider both cases.

\textbf{Case 1 (siblings).} Let $p$ denote their common parent. For any $k$ outside the subtree rooted at $p$, the additivity property gives
\[
d_{\ell}(\mathbf{z}_i, \mathbf{z}_k) = d_{\ell}(\mathbf{z}_i, \mathbf{z}_p) + d_{\ell}(\mathbf{z}_p, \mathbf{z}_k), \quad
d_{\ell}(\mathbf{z}_j, \mathbf{z}_k) = d_{\ell}(\mathbf{z}_j, \mathbf{z}_p) + d_{\ell}(\mathbf{z}_p, \mathbf{z}_k).
\]
Subtracting,
\[
d_{\ell}(\mathbf{z}_i, \mathbf{z}_k) - d_{\ell}(\mathbf{z}_j, \mathbf{z}_k) 
= d_{\ell}(\mathbf{z}_i, \mathbf{z}_p) - d_{\ell}(\mathbf{z}_j, \mathbf{z}_p),
\]
which is constant across $k$. Therefore $D_{ij}=0$.

\textbf{Case 2 (parent–child).} Suppose $i$ is the parent of $j$. For any $k$ outside the subtree rooted at $i$,
\[
d_{\ell}(\mathbf{z}_j, \mathbf{z}_k) = d_{\ell}(\mathbf{z}_j, \mathbf{z}_i) + d_{\ell}(\mathbf{z}_i, \mathbf{z}_k),
\]
so
\[
d_{\ell}(\mathbf{z}_i, \mathbf{z}_k) - d_{\ell}(\mathbf{z}_j, \mathbf{z}_k) = - d_{\ell}(\mathbf{z}_j, \mathbf{z}_i),
\]
a constant in $k$. Hence $D_{ij}=0$ also in this case.

\paragraph{Converse direction ($D_{ij}=0 \Rightarrow$ same bottom set).}
If $D_{ij}=0$, then
\[
d_{\ell}(\mathbf{z}_i,\mathbf{z}_k) - d_{\ell}(\mathbf{z}_j,\mathbf{z}_k)
\]
is constant for all $k \in \mathcal{V}\setminus\{i,j\}$. Let $p$ denote the nearest common ancestor of $i$ and $j$. For any node $k$ not a descendant of $p$, additivity implies
\[
d_{\ell}(\mathbf{z}_i,\mathbf{z}_k) = d_{\ell}(\mathbf{z}_i,\mathbf{z}_p) + d_{\ell}(\mathbf{z}_p,\mathbf{z}_k), \quad
d_{\ell}(\mathbf{z}_j,\mathbf{z}_k) = d_{\ell}(\mathbf{z}_j,\mathbf{z}_p) + d_{\ell}(\mathbf{z}_p,\mathbf{z}_k).
\]
Subtracting,
\[
d_{\ell}(\mathbf{z}_i,\mathbf{z}_k)-d_{\ell}(\mathbf{z}_j,\mathbf{z}_k)
= d_{\ell}(\mathbf{z}_i,\mathbf{z}_p)-d_{\ell}(\mathbf{z}_j,\mathbf{z}_p),
\]
which is constant in $k$. This implies both $i$ and $j$ must attach directly to $p$, either as siblings or in a parent–child relation. Hence $i$ and $j$ belong to the same bottom set.

\paragraph{Practical note.}
In real data, due to sampling noise and embedding estimation error, $D_{ij}$ will rarely be exactly zero even for nodes within the same bottom set. Our algorithm therefore computes the full matrix $[D_{ij}]_{1 \le i,j \le |\mathcal{V}|}$ and applies $k$-means clustering. The silhouette method is used to select the number of clusters, providing a practical approximation to the theoretical bottom set criterion.
\end{document}